\def\year{2020}\relax
\documentclass[letterpaper]{article} 
\usepackage{aaai20}  
\usepackage{times}  
\usepackage{helvet} 
\usepackage{courier}  
\usepackage[hyphens]{url}  
\usepackage{graphicx} 
\usepackage{amsmath, amssymb}
\usepackage{comment}
\usepackage{booktabs}
\usepackage{color}
\usepackage{graphicx}  
\title{Combination of Unified Embedding Model\\ and Observed Features\\ for Knowledge Graph Completion}

\author{Takuma Ebisu${}^\text{1,2}$\and Ryutaro Ichise${}^\text{2,1,3}$\\
	${}^\text{1}$SOKENDAI (The Graduate University for Advanced Studies)\\
	2-1-2 Hitotsubashi, Chiyoda-ku, Tokyo, Japan\\
	${}^\text{2}$National Institute of Informatics\\
	2-1-2 Hitotsubashi, Chiyoda-ku, Tokyo, Japan\\
	${}^\text{3}$National Institute of Advanced Industrial Science and Technology\\
	2-3-26 Aomi, Koto-ku, Tokyo, Japan\\
	\{takuma,ichise\}@nii.ac.jp\\
}

\begin{document}

\maketitle

\begin{abstract}
Knowledge graphs are useful for many artificial intelligence tasks but often have missing data.
Hence, a method for completing knowledge graphs is required.
Existing approaches include embedding models, the Path Ranking Algorithm, and rule evaluation models.
However, these approaches have limitations.
For example, all the information is mixed and difficult to interpret in embedding models, and traditional rule evaluation models are basically slow.
In this paper, we provide an integrated view of various approaches and combine them to compensate for their limitations.
We first unify state-of-the-art embedding models, such as ComplEx and TorusE, reinterpreting them as a variant of translation-based models.
Then, we show that these models utilize paths for link prediction and propose a method for evaluating rules based on this idea.
Finally, we combine an embedding model and observed feature models to predict missing triples. This is possible because all of these models utilize paths.
We also conduct experiments, including link prediction tasks, with standard datasets to evaluate our method and framework.
The experiments show that our method can evaluate rules faster than traditional methods and that our framework outperforms state-of-the-art models in terms of link prediction.
\end{abstract}

\section{Introduction}
Knowledge graphs are used to describe many types of real-world relations in a form that can be easily processed by a computer. 
Several knowledge graphs, such as YAGO \cite{DBLP:conf/www/SuchanekKW07}, DBpedia \cite{DBLP:conf/semweb/AuerBKLCI07}, and Freebase \cite{Bollacker:2008:FCC:1376616.1376746}, have been recently developed {and applied for many artificial intelligence tasks \cite{Hakimov:2012:NER:2237867.2237871,Daiber:2013:IEA:2506182.2506198,D14-1067}.}
These knowledge graphs can never be complete because the numbers of entities and relations are huge and new entities and relations are frequently created while human resources are limited.
Hence, a system is needed for predicting missing data to automatically complete knowledge graphs.

In a knowledge graph, a fact is represented by a labeled and directed edge, called a triple $(h,r,t)$, where $h$ and $t$ are entity nodes and $r$ is the relation label of an edge from $h$ to $t$.
Many kinds of model for link prediction  have been developed to estimate unknown facts, where link prediction is the task of predicting an entity to answer a query, i.e., a triple with a missing value, such as $(h,r,?)$ or $(?,r,t)$.
Several approaches have been proposed for link prediction, such as knowledge graph embedding models, rule evaluation models, and the Path Ranking Algorithm (PRA).
Although these models were independently developed, they all use paths, as discussed in the following sections.

In this paper, we integrate these approaches and make them efficiently work together. The main contributions in this paper are as follows:
\begin{itemize}
	\item We unify state-of-the-art knowledge graph embedding models and find a connection between rules and embedding models.
	\item We propose a method that evaluate rules based on embeddings.
	\item We propose a framework for combining approaches for link prediction to compensate for the disadvantages of each approach.
	\item We evaluate the proposed method and framework in terms of calculation time and link prediction accuracy with standard datasets. It is shown that our method can find useful rules faster than traditional rule evaluation models and that our framework outperforms other models in terms of link prediction.
\end{itemize}

The remainder of this paper is organized as follows.
In Section \ref{relatedwork}, we discuss related work on link prediction.
In Section \ref{trans-bi}, we unify state-of-the-art embedding models and formally discuss their utilization of path information for link prediction.
In Section \ref{e_rule}, we propose a method for evaluating and selecting useful rules for link prediction based on embeddings.
In Section \ref{framework}, we propose a framework for combining many approaches.
In Section \ref{experiments}, we present an experimental study that compares our method and framework with baseline results for benchmark datasets.
In Section \ref{conclusions}, we present the conclusions.
\section{Related Work}
\label{relatedwork}
A number of models have been developed based on various approaches. We divide the main approaches for link prediction into two groups, namely embedding models and observed feature models.
We summarize these approaches and describe their advantages and disadvantages. It is shown that they are complementary to each other for link prediction.

The following notation is used to discuss related work and our work. 
$e$ and $r$ denote an entity and a relation of a knowledge graph.
$E$ and $R$ respectively represent sets of entities and relations.
Then, a knowledge graph is described as a set of triples: $KG= \{(h,r,t)\}\in E\times R\times E$, where $h$, $r$, and $t$ are called the head entity, relation, and tail entity, respectively. 
We add the inverse relation $r^{-1}$ to $R$ for each $r\in R$ and add the inverse triple $(t,r^{-1},h)$ to $KG$ for each $(h,r,t)\in KG$ to facilitate explanation.

\subsection{Knowledge Graph Embedding Models}
Knowledge graph embedding models embed entities and relations in a vector space.
We mainly discuss translation-based models and bilinear models because they are very simple and have been shown to be more efficient than more complex models such as neural-network-based models \cite{DBLP:journals/corr/KipfW16,DBLP:journals/corr/DettmersMSR17}.

In a conventional translation-based model, a link between two entities is represented by a certain translation operation on the embedding space. 
This is formally described by the principle $\boldsymbol{h}+\boldsymbol{r}= \boldsymbol{t}$, where $\boldsymbol{h},\boldsymbol{r}$, and $\boldsymbol{t}$ are the embeddings of $h$, $r$, and $t$, respectively.
The first translation-based model was TransE \cite{DBLP:conf/nips/BordesUGWY13}, which embeds entities and relations in a real vector space.
However, the conflict between the principle and regularization is problematic. TorusE \cite{DBLP:journals/corr/abs-1711-05435} was proposed to solve this problem by changing the embedding space to a torus manifold.
A more generalized concept called Knowledge Graph Embedding on a Lie Group (KGLG) \cite{KGLG} has been recently proposed.
Here, TransE and TorusE are interpreted as instances of KGLG.
We believe that KGLG efficiently captures first-order rules for link prediction.
More complex models, such as TransR \cite{DBLP:conf/aaai/LinLSLZ15} and TransD \cite{ji-etal-2015-knowledge}, have been proposed that have more degrees of freedom by mapping embeddings to other spaces depending on the relation to overcome the low expressiveness of TransE.
However, these models have not been shown to be clearly effective because their high expressivity makes it difficult to capture first-order rules.
TransH \cite{DBLP:conf/aaai/WangZFC14} and TransAt \cite{ijcai2018-596} select a subspace depending on the relation when the principle is applied.
These models are discussed in detail in Section \ref{low_expressiveness}. We propose generalized KGLG in Section \ref{AKGLG} that retains the ability to make use of rules.
TransH and TransAt can be considered as a restricted version of generalized KGLG.

Bilinear models represent a relation as a bilinear function and treat the embeddings of entities as arguments of that function to score a triple. 
RESCAL \cite{DBLP:conf/icml/NickelTK11}, the first bilinear model, represents
each relation as a bilinear function.
RESCAL is the most general form of a bilinear model. 
Hence, it tends to overfit training data.
Extensions of RESCAL have been proposed by restricting the bilinear functions. For example,
DistMult \cite{DBLP:journals/corr/YangYHGD14a} and ComplEx \cite{DBLP:conf/icml/TrouillonWRGB16} restrict the matrices representing the relations to diagonal matrices.
We show that these models can be considered as extended KGLG, as discussed in Section \ref{trans-bi}.

{The main problems with existing embedding models are the lack of interpretability (i.e., the models do not give a reason for a prediction) and the mixing of all information in embeddings even though a certain relation may require only a few simple rules to precisely predict.}

\subsection{Observed Feature Models}
Observed feature models directly utilize observed features. They can be divided into rule evaluation models and PRA.
The main advantage of these models over knowledge graph embedding models is their interpretability and information selectivity. 
Hence, they overcome the problems of embedding models.

AMIE \cite{DBLP:conf/www/GalarragaTHS13,DBLP:journals/vldb/GalarragaTHS15} is a well-known model that evaluates and extracts the rules underlying a knowledge graph.
It has several problems, including imbalance of the partially complete assumption \cite{ebisu-ichise-2019-graph}. 
GRank \cite{ebisu-ichise-2019-graph} was proposed to deal with these problems; its performance is competitive with that of embedding models.
However, there are still some problems, such as slow calculation speed and non-integrated rules.
Another problem is the limited search space.
Practically, the rules are limited to the form $r_1(x_1,x_2)\wedge r_2(x_2,x_3)\wedge \cdots \wedge r_n(x_n,x_{n+1})\Rightarrow r(x_1,x_{n+1})$, where $x_i$ is a variable; hereafter, we refer to a rule as $(r_1,r_2,...,r_n)\Rightarrow r$.
More complex rules or those that include constants are not considered because such rules are often useless and they greatly expand the search space.

PRA \cite{DBLP:journals/ml/LaoC10,DBLP:conf/emnlp/LaoMC11} constructs logistic classification models for each relation based on features that represent the existence of a particular path between two entities.
However, the models lack an efficient way to select paths for features.
This problem can be overcome using rule evaluation models.

The main problems of observed feature models are slowness and a limited rule search space.
To solve these problems, we propose a method for evaluating rules based on embeddings in Section \ref{Rule_AKGLG} and a framework for combining embedding models and observed feature models in Section \ref{framework}.
Some models \cite{DBLP:conf/emnlp/GuoWWWG16,DBLP:conf/aaai/GuoWWWG18} employ rules extracted by traditional rule evaluation models to obtain better embeddings. In contrast, we refine the information in embeddings based on the observed features. This is done because embedding models already can capture rules but cannot order information.

\section{Unification of Knowledge Graph Embedding Models}
\label{trans-bi}

The concept of KGLG allows us to take any Lie group as the embedding space of a translation-based model.
KGLG solves the problem caused by regularization. However, another problem still remains, as discussed in the following section. 
We propose a concept of embeddings called Attentioned Knowledge Graph Embedding on Lie Group (AKGLG) to solve this problem by generalizing KGLG.
We show that state-of-the-art embedding models are instances of AKGLG.
\subsection{Mechanism of Translation-based Models}
\label{low_expressiveness}
In KGLG, relations and entities are represented by points on a Lie group $G$ following the generalized principle ${g_{h}}+_G g_{r}= g_{t}$, where $+_G$ is the group operation of $G$ and $g_{h}$, $g_{r}$, and $g_{t}$ are embeddings on $G$ of the head entity, relation, and tail entity, respectively, of an observed triple $(h,r,t)$.
$G$ has a similarity function $d_{G}$ that is used to score a triple $(h,r,t)$ with $d_{G}({g_{h}}+_G g_{r}, g_{t})$.
This principle allows KGLG to utilize first-order rules based on a path.
If a rule $(r_1,r_2,...,r_n)\Rightarrow r$ holds and there are enough groundings , i.e., a mapping from variables in the rule to $E$ holding relations \cite{ebisu-ichise-2019-graph}, in a knowledge graph, then the embeddings of these relations are trained to follow the equation: 
\begin{eqnarray}
\sum_{k=1}^{i}g_{r_k}:=g_{r_1}+_{G}g_{r_2}+_{G}\cdots +_{G}g_{r_k}=g_{r}
\end{eqnarray}
We can get this equation by the sequential application of the principle.
However, the principle seems too strict to compatibly embed various entities and relations.
For example, if a head entity/relation pair have multiple valid tail entities and the embeddings perfectly follow the principle, then all of the tail entities have to be represented by the same point; this is undesirable because we need to distinguish different entities.
TransR and TransD solve this problem by mapping entities to another space depending on the relation when the principle is applied, where the embeddings of relations are on codomains of these mappings.
However, these models cannot utilize rules because the embeddings of relations are in different spaces and thus equation (1) has no meaning.
Hence, we need to extend KGLG in a different way.
\subsection{Attentioned Knowledge Graph Embedding on Lie Group}
\label{AKGLG}
The problem discussed in the previous section occurs because entities and relations are equally distributed throughout the embedding space of KGLG.
We solve the problem by assigning an attention vector for each entity and relation {to structuralize KGLG}.
The attention vector indicates the part of the embedding space where the information of the corresponding entity or relation is stored.
We construct AKGLG on KGLG, whose embedding space is denoted by $G$.
For AKGLG, entities $e$ and relations $r$ are represented by points $\boldsymbol{g}_e$ and $\boldsymbol{g}_r$, respectively, on $G'=G^n$.
Then, we assign vectors $w_e$ and $w_r \in [0,\infty)^n $ to each entity and each relation, respectively.
The score of AKGLG of a triple $(h,r,t)$ is formally defined as follows:
\begin{displaymath}
Score_{G'}({h}, {r},{t})=(w_h\circ w_r \circ w_t)\cdot \boldsymbol{d}_{G}(\boldsymbol{g}_h\boldsymbol{+}_{G} \boldsymbol{g}_r,\boldsymbol{g}_t)
\end{displaymath}
where $\circ$ represents the element-wise multiplication operation, $\cdot$ represents the dot product operation, and $\boldsymbol{d}_{G}(\boldsymbol{g}_h\boldsymbol{+}_{G} \boldsymbol{g}_r,\boldsymbol{g}_t)$ is an $n$-dimensional  vector whose $i$-th  element is equal to ${d}_{G}({h_i}+_{G}{r_i},{t_i})$.
Note that when the score of $(h,r,t)$ is calculated, only the part of $G'$ where the attentions of $h$, $r$, and $t$ overlap, i.e., their attention values are simultaneously large enough, is considered.
These attentions and the score function produce embeddings of entities and relations without conflict by properly separating the stored information.
\subsection{Existing Embedding Models as Instances of AKGLG}
\label{instances}
\begin{table}[t]
	\caption{List of existing embedding models that can be interpreted as instances of KGLG and AKGLG.
	Note that TransAt is more restricted than standard AKGLG.}\smallskip
	\centering
	\resizebox{.85\columnwidth}{!}{
		\smallskip\begin{tabular}{c|l|l}\hline
			Base Lie Group & KGLG & AKGLG\\\hline
			\{-1,1\} & Not proposed & DistMult\\
			$\mathbb{R}$ & TransE & TransH, TransAt\\
			$S^1$& TorusE & ComplEx\\\hline
		\end{tabular}
	}
	\label{lists of KGLG}
\end{table}
Examples of AKGLG and KGLG are shown in Table \ref{lists of KGLG}.

We can consider a KGLG on a group $G_1=\{-1,1\}$, where the group operation is the standard real number multiplication operation.
We define the similarity function on $G_1$ as $d_1(x,y)=1 (\text{if}\, y=x),-1 (\text{if}\,x\neq y)$.
Then, we can consider a KGLG on $G_1^n$ that extends the similarity function of $G_1$, i.e., it takes the sum after the element-wise calculation.
$G_1^n$ is not an infinite set but can utilize simple rules.
An AKGLG on $G_1^n$ is equivalent to DistMult.
Here, each entity or relation has its attention vector and its embedding on $G_1^n$.
We can obtain a vector representation for each relation or entity element-wisely multiplying the attention vector and the embedding as real vector. The triple score of a DistMult based on these vector representations is the same as the score of AKGLG, i.e.:
\begin{eqnarray*}
Score_{G_1^n}(h,r,t)=(w_h\circ w_r \circ w_t)\cdot \boldsymbol{d}_{G_1}(\boldsymbol{g}_h\boldsymbol{+}_{G_1} \boldsymbol{g}_r,\boldsymbol{g}_t)
\\=(w_h\circ\boldsymbol{g}_h)\circ ( w_r \circ \boldsymbol{g}_r) \cdot (w_t\circ \boldsymbol{g}_t)
\end{eqnarray*}
where the right side is the score of DistMult.

We can consider a KGLG on a circle $S^1$ as a subset of $\mathbb{C}$ whose elements have  a magnitude of 1 where the group operation $+_{S^1}$ is the standard complex number multiplication operation.
We define the similarity function as $d_{S^1}(x,y)=Re(x+_{S^1}\overline{y})$.
Then, we can consider a KGLG on $S^n$ that extends the score function of $S^1$; this is equivalent to TorusE \cite{DBLP:journals/corr/abs-1711-05435}.
An AKGLG on $S^n$ is equivalent to ComplEx.
We can obtain a complex vector representation for each relation or entity multiplying them . The triple score of ComplEx based on these vector representations is the same as the score of AKGLG, i.e.:
\begin{eqnarray*}
	Score_{S^n}(h,r,t)=(w_h\circ w_r \circ w_t)\cdot \boldsymbol{d}_{S^1}(\boldsymbol{g}_h\boldsymbol{+}_{S^1} \boldsymbol{g}_r,\overline{\boldsymbol{g}_t})
	\\=Re((w_h\circ\boldsymbol{g}_h)\circ ( w_r \circ \boldsymbol{g}_r) \cdot (w_t\circ \overline{\boldsymbol{g}_t}))
\end{eqnarray*}
where the right side is the score of ComplEx.

We can consider a KGLG on $\mathbb{R}$ where the group operation is the standard summation operation.
We define the similarity function as $d_\mathbb{R}(x,y)=-(x-y)^2$.
Then, we can consider a KGLG on $\mathbb{R}^n$ that extends the score function of $\mathbb{R}$; this is equivalent to TransE.
An AKGLG on $\mathbb{R}^n$ has not been proposed.
However, its restricted versions are TransH and TransAt.
In TransAt, the attention vectors for entities are fixed to the vector whose elements are all 1 and the attention vectors for relations are restricted to a vector whose elements are 0 or 1.

As we have shown, knowledge graph embedding models can be unified using the concept of AKGLG. These models work based on the translation principle.

\section{Rule Evaluation on Embeddings}
\label{e_rule}
In the previous section, we proposed AKGLG.
We showed that both KGLG and AKGLG utilize rules.
In this section, we propose a method for evaluating rules based on this idea.
{This method allows us to interpret embeddings; that is, we can know what kind of rules are learned and used for link prediction.
Additionally, the method evaluates rules faster than traditional rule evaluation methods.}

\subsection{Rule Evaluation on KGLG}
Ebisu et al. \shortcite{ebisu-ichise-2019-graph} proposed a method that assigns two confidence scores to a rule, where one is the confidence of predicting a tail entity and the other is that of predicting a head entity.
Hence, we assume that the rule $(r_1,r_2,...,r_n)\Rightarrow r$ is used to predict a tail entity given a head entity.
The rule for predicting a head entity of $r$ based on the same path is described by its inverse form: $(r_n^{-1},r_{n-1}^{-1},...,r_1^{-1})\Rightarrow r^{-1}$.
We also suppose that a KGLG learns two embeddings for each relation, as proposed by Lacroix et al. \shortcite{pmlr-v80-lacroix18a}.
One of the embeddings is trained and used to predict the tail entities of the corresponding relation; we denote this embedding as $g_r$.
The other is trained and used to predict head entities; we denote this embedding as $g_{r^{-1}}$.

If $(r_1,r_2,...,r_n)\Rightarrow r$ is really useful for predicting tail entities related with $r$, then the KGLG learns embeddings in such a way that $g_{(r_1,r_2,...,r_n)}=\sum_{k=1}^{i}g_{r_k}$ and $g_{r}$ are similar.
Therefore, we propose to evaluate the rule by measuring the similarity between $g_{(r_1,r_2,...,r_n)}$ and $g_{r}$.
KGLG has similarity function $d_{G}$, which is used to estimate the validity of a triple, as discussed in Section \ref{instances}.
For example, TransE has L1 norm or the square of L2 norm.
We can use the function to measure the similarity between $g_{(r_1,r_2,...,r_n)}$ and $g_{r}$.
The confidence score of a rule $(r_1,r_2,...,r_n)\Rightarrow r$ is formally written as follows:
\begin{displaymath}
conf((r_1,r_2,...,r_n)\Rightarrow r)=d_G(g_{(r_1,r_2,...,r_n)},g_{r})
\end{displaymath}
However, the confidence score obtained using this equation is not reliable because the principle is too restrict to obtain good embeddings.
In the following section, we define the confidence score on AKGLG to evaluate a rule more properly.
\subsection{Rule Evaluation on AKGLG}
\label{Rule_AKGLG}
\subsubsection{Path Representation}
For AKGLG, each entity and relation is represented by its embedding on a Lie group and its attention vectors.
We want to represent a path to compare it with the representation of a relation and evaluate a rule on AKGLG.
For the embeddings on a Lie group, we can obtain the path embedding on a Lie group by group multiplying the relation embeddings on the path, as we did in the previous section.
For the attention vectors, we take the element-wise geometric mean of the attention vectors on the path.
We employ the geometric mean because we want the $i$-th element to be zero if one of the $i$-th elements of the attention vectors is 0 because the information does not propagate through the corresponding dimension.
We then normalize the path attention vector because the magnitude of attention vectors likely represents  the appearance frequency in the knowledge graph; we thus do not want to take the magnitude into account.
The attention vector of a path $(r_1,r_2,...,r_n)$ is formally defined as follows:
\begin{displaymath}
w_{(r_1,r_2,...,r_n)}=\frac{(w_{r_1}\circ w_{r_2}\circ \cdots \circ w_{r_n})^{1/n}}{||(w_{r_1}\circ w_{r_2}\circ \cdots \circ w_{r_n})^{1/n}||_2}
\end{displaymath}
\subsubsection{Evaluation of Rules}
The similarity of the embeddings on a Lie group is calculated in the same way as that for KGLG.
We take the attention vectors into account in a way similar to that used for calculating the score of a triple.
The confidence score of a rule $(r_1,r_2,...,r_n)\Rightarrow r$ on AKGLG is formally defined as follows:
\begin{eqnarray*}
\begin{split}
\it{conf}&((r_1,r_2,...,r_n)\Rightarrow r)
\\&=(w_{(r_1,r_2,...,r_n)}\circ w_{r})\cdot \boldsymbol{d}_{G}(\boldsymbol{g}_{(r_1,r_2,...,r_n)},\boldsymbol{g}_{r})
\end{split}
\end{eqnarray*}
where $\boldsymbol{d}_{G}(\boldsymbol{g}_{(r_1,r_2,...,r_n)},\boldsymbol{g}_{r})$ is the $n$-dimensional vector whose $i$-th element is equal to $d_{G}({g_{(r_1,r_2,...,r_n),i}}, {g}_{r,i})$.
{We refer to this evaluation method as the rule evaluation based on embeddings (REE).}
\section{Framework for Combining Link Prediction Approaches}
\label{framework}
\begin{figure}[tb]
	\centering
	\includegraphics[scale=0.45]{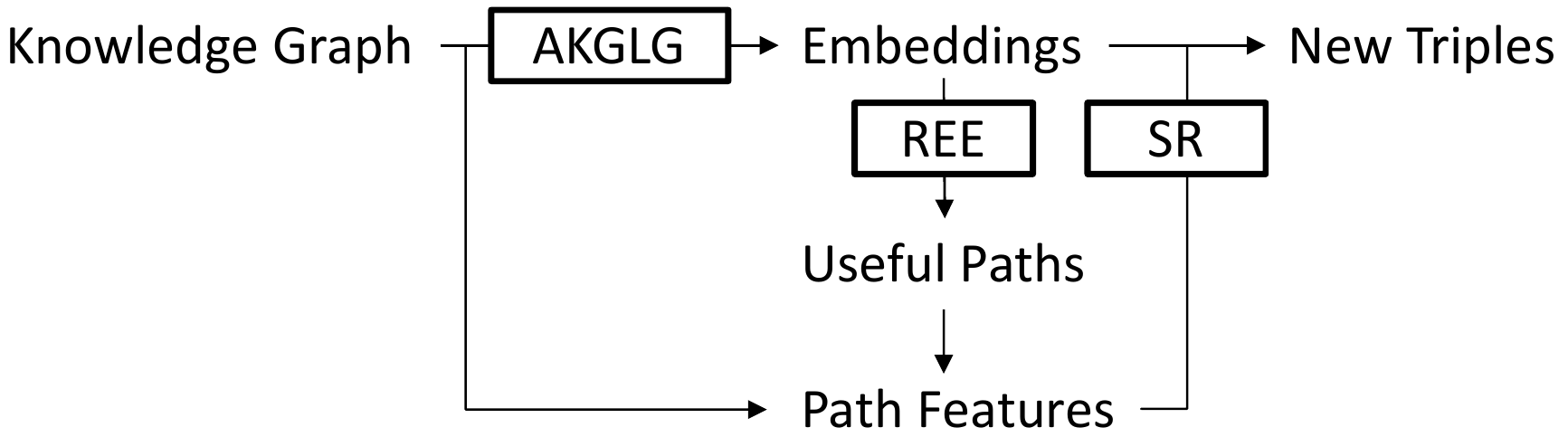}
	\caption{Procedure of PBF. In this figure, SR stands for softmax regression.}
	\label{flow}
\end{figure}
{In this section, we propose a framework to exploit the advantages of various approaches for link prediction, as discussed in Section \ref{relatedwork}.}
The outline of the framework is as follows: 
\begin{enumerate}
	\item Obtain embeddings of entities and relations by employing AKGLG.
	\item Extract useful paths for each relation for link prediction by evaluating corresponding rules with traditional rule evaluation models or REE.
	\item Construct a softmax regression model for each relation in a way similar to PRA. The features used for training and prediction are obtained by counting the number of groundings of extracted paths, as done by GRank.
	\item Perform link prediction by taking the weighted sum of the scores of the embedding model  and the softmax regression models. 
\end{enumerate}
{We refer to this framework as the path-based framework (PBF).}
The flowchart of PBF with REE is shown in Figure \ref{flow}.
We discuss the details of steps 2 and 3 below.
\subsection{Path Extraction using REE}
{
We can select useful paths for each relation using traditional rule evaluation models.
However, traditional models are time-consuming because the number of candidate paths increases exponentially with their size, and so do their groundings.
We expect that REE can evaluate rules faster because it does not need to consider groundings.
}

We first extract candidate rules by finding positive groundings for REE.
This can be done much faster than evaluating rules with traditional methods because we do not need to find negative groundings of rules.
We restrict groundings to ``injective" , i.e., one entity can appear at most once, as proposed by \citeauthor{ebisu-ichise-2019-graph} \shortcite{ebisu-ichise-2019-graph}.
This restriction allow us to find groundings on the converted simple graph (i.e., there are no multiple edges between entities) using the following procedure.
In the explanation, we limit the path length to be an odd number $2k-1$  for simplicity and we suppose that the entities are ordered (i.e., labeled by different integers). 
\begin{enumerate}
	\item Convert a knowledge graph $KG = \{(h,r,t)\}$ to the simple graph $\{(h,r)|(h,r,t)\in KG\}$.
	\item For each entity $e$, find all cycles in which $e$ is the smallest entity as follows:
	\begin{itemize}
		\item Find all entity paths whose length is at most $k$ under the condition that all entities except $e$ on paths are larger than $e$ in terms of the order.
		\item For each pair of entity paths whose last entities are the same, make a cycle that concatenates them and ensure that there is no entity duplication.
	\end{itemize}
	\item For each path $p$, if there is a cycle that is the groundings  of the rule $p \Rightarrow r$, then add the rule to the candidate rules to be evaluated by REE.
\end{enumerate}
Note that we can find cycles with no duplication and conduct the computations using parallel processing  because the procedure for each entity is independent.
Then, we evaluate candidate rules with REE and select a fixed number of rules for each relation.
{The body paths of these rules are used to construct softmax regression models.}
\subsection{Softmax Regression Model}
In this section, we construct a softmax regression model for each relation following PRA.
We first construct training queries for each relation.
Training queries for relation $r$ are formally defined as $\{(h,r,?)|(h,r,t)\in KG\}$.
Each element of the feature vector of a query $(h,r,?)$ and an entity $e$ corresponds to one of the extracted paths and its value is equal to the multiplicity of the corresponding path (i.e., the number of groundings of the path that start from $h$ and end at $e$).
We employ multiplicity because it is efficient, as shown by GRank.
However, the magnitude of path multiplicity greatly differs.
Hence, we need to rescale feature vectors to obtain a good model. 
The $i$-th element of the final feature vector for a query $(h,r,?)$ and an entity $e$ is formally defined as follows:
\begin{eqnarray*}
	v_{((h,r,?),e),i}=mul(h,e,p_i)/M
\end{eqnarray*}
where $p_i$ is the path indexed by $i$ for $r$, $mul(h,e,p_i)$ is the number of groundings of $p_i$ that start from $h$ and end at $e$, which is equivalent to $score(p_i,(h,e))$ defined for GRank \cite{ebisu-ichise-2019-graph}, and $M=max\{mul(h,e',p_i)|e'\in E\}$. 

{
Next, we describe the softmax regression models.
The softmax regression model for $r$ has a parameter vector $\theta_r$ for training. The score of the triple $(h,r,e)$ for query $(h,r,?)$ is calculated by taking the dot product of $\theta_r$ and $v_{((h,r,?),e)}$.
For each answer for a query, a fixed number of entities are randomly selected for negative examples.
We employ cross entropy for the loss function.
The loss function for the model of a relation $r$ is formally written as follows:
\begin{eqnarray*}
	L_r=\sum_{(h,r,t)\in KG}{-log(\frac{exp(\theta_r^\mathsf{T}\cdot v_{((h,r,?),t)})}{\sum_{e\in NE_{(h,r,t)} \cup \{t\}}{exp(\theta_r^\mathsf{T}\cdot v_{((h,r,?),e)})}})}
\end{eqnarray*}
}
{
where $NE_{(h,r,t)}$ is the set of randomly selected negative entities.
Note that $v_{((h,r,?),e)}$ can often be the zero vector. 
We ignore the positive triple $(h,r,t)$ if $v_{((h,r,?),t)}$ is the zero vector and we do not select the entity $e$ as a negative example if $v_{((h,r,?),e)}$ is the zero vector.
}
\section{Experiments}
\label{experiments}
In this section, we conduct experiments to evaluate REE and PBF.
REE is directly compared with traditional rule evaluation models in terms of calculation time and link prediction accuracy.
PBF is compared with other models in terms of link prediction accuracy. 
\subsection{Datasets}
\begin{table}[tb]
	
	\caption{Statistics of benchmark datasets.}\smallskip 
	
	\scalebox{0.80}{
		\begin{tabular}{c|cccc}
			&WN18&WN18RR&FB15k&FB15k-237\\\hline 
			\# of Entities&40,943&40,943&14,951&14,541\\
			\# of Relations&18&11&1,345&237\\
			\# of Training Triples &141,442&86,835&483,142&272,115\\
			\# of Validation Triples &5,000&3,034&50,000&17,535\\
			\# of Test Triples&5,000&3,134&59,071&20,466\\

		\end{tabular}
	}
	\label{Table 2}
\end{table}
Experiments were conducted on four benchmark datasets, namely WN18, FB15k \cite{DBLP:conf/nips/BordesUGWY13}, WN18RR \cite{DBLP:journals/corr/DettmersMSR17}, and FB15k-237 \cite{observed-versus-latent-features-for-knowledge-base-and-text-inference} (details are shown in Table \ref{Table 2}).
These datasets have been widely used for evaluating model performance in link prediction tasks.

WN18 and FB15k are extracted from the real knowledge graphs WordNet \cite{Miller:1995:WLD:219717.219748} and Freebase \cite{Bollacker:2008:FCC:1376616.1376746}, respectively.
WordNet is a well-known human-curated lexical database and
Freebase is a huge knowledge graph of general facts, but has many missing facts.
WN18 and FB15k have redundancy in the form of reverse relations.  
When WN18RR and FB15k-237 are extracted from WN18 and FB15k, these inverse relations are removed.

\subsection{Protocol of Link Prediction Task}
We conducted a link prediction task following the approach reported by \citeauthor{DBLP:conf/nips/BordesUGWY13} \shortcite{DBLP:conf/nips/BordesUGWY13} to evaluate our methods.
For each test triple $(h_t,r_t,t_t)$ in a dataset, two queries, $(h_t,r_t,?)$ and $(?,r_t,t_t)$, were constructed.
Then, we obtained the rankings of entities for each query using each method, as outlined below.
The rankings were filtered by eliminating entities whose corresponding triples (except the target test triple) were included in the training, validation, or test triples .
The obtained rankings were scored in terms of the mean reciprocal rank (MRR) and HITS@\textit{n},
where MRR is the mean of the inverse of the ranks of the corresponding entities and HITS@\textit{n} is the proportion of test queries whose corresponding entities are ranked in the top \textit{n} of the obtained rankings.

Next, we describe how to obtain rankings using the methods. 
For PBF, we can get the rankings of entities for a query $(h_t,r_t,?)$ by calculating the score of triples $(h_t,r_t,e)$ for each $e\in E$.
For REE, we follow the settings of \citeauthor{ebisu-ichise-2019-graph} \shortcite{ebisu-ichise-2019-graph} to obtain ranking entities from the extracted rules.
We extract 1,000 rules for each relation (including the inverse of a relation) and each of the rules is used to obtain entity rankings for a query by counting its groundings.
The final rankings of entities are obtained by concatenating the rankings from each rule.
\subsection{Experimental Settings}
We employed ComplEx in the experiments as an instance of AKGLG. ComplEx was used to obtain embeddings to extract rules in the first experiment and answer test queries in the second experiment.
Note that each entity and relation is represented by a complex vector $\boldsymbol{c}\in \mathbb{C}^n$. This vector is decomposed into attention vector $\boldsymbol{abs}(\boldsymbol{c})$ and point on the torus $(\boldsymbol{1}/\boldsymbol{abs}(\boldsymbol{c}))\circ \boldsymbol{c}$, where $\boldsymbol{abs}(\boldsymbol{c})$ is a real vector whose $i$-th element is equal to $abs(c_i)$.
The settings and hyperparameters of ComplEx are those given by Lacroix et al. \shortcite{pmlr-v80-lacroix18a}, with a dimension of 2,000 and L3 regularization.
We also conducted experiments on DistMult and TorusE with the same settings for fair comparison.

We employed GRank with fdMAP \cite{ebisu-ichise-2019-graph} for comparison with REE and for use in PBF.
The limit of the path length for rules was selected from $\{1,2,3\}$ based on the MRR of link prediction on the validation triples.

For the softmax regression models in PBF, we set the number of extracted paths for each relation to 100.
We employed stochastic gradient descent for training.
We used $L_2$ regularization and the coefficient of the regularization. The learning rate was selected from $\{0.1,0.01,0.001\}$ depending on the MRR of link prediction on the validation data . 
We set batch size to 100 and trained each model using 500 batches.

The weight for the final step of PBF was selected from $\{0,0.1,\cdots,1\}$ based on the MRR of link prediction on the validation data .

\subsection{Experimental Results for REE}
\subsubsection{Calculation Time}
Table \ref{Table 3} shows the calculation times for REE, including the rule candidate selection method (see Section 5.1), and GRank for the relatively large datasets FB15k and FB15k-237. It took GRank less than one minute to finish for WN and WN18RR.
Note that we used an Intel Xeon Gold 6140 CPU (18 cores) for running GRank and rule candidate selection and an Intel Xeon E5-1620 CPU (4 cores) and a GPU (Nvidia Titan X) for running REE.
The maximum path size was 3 in this experiment.
The results show that REE is more efficient than GRank, especially for FB15k.
FB15k is a denser graph than FB15k-237 and thus the number of groundings of rules is larger.
That makes GRank even slower.
As a result, REE is 30 times faster than GRank.

We did not take the computation time of ComplEx, which is relatively short, into account.
Lacroix et al. \shortcite{pmlr-v80-lacroix18a} reported that one epoch of training for ComplEx on FB15k takes about 110 s and that 25 epochs are sufficient.
For hyperparameter tuning, one epoch is sufficient.
Hence, about 5,000 s are sufficient to obtain embeddings; this is far faster than rule evaluation using GRank.
To estimate rules faster, we can employ low-dimensional space, as described in Section \ref{less dimension}. 
\begin{table}[tb]
	\caption{Calculation time of REE and GRank.}\smallskip
	\centering
	\scalebox{0.85}{
		\begin{tabular}{c|cc}
			
			&FB15k&FB15k-237\\\hline \hline
			GRank&181,352s&2,486s\\
			REE&5,504s&544s\\

	\end{tabular}}
	
	\label{Table 3}
\end{table}
\subsubsection{Link Prediction Task}
	The results of the link prediction tasks for REE are shown in Tables \ref{Table 4}. The results reported in previous studies are included for comparison.
	We focus on the comparison of REE with traditional rule evaluation models GRank and GPro \cite{ebisu-ichise-2019-graph}, which is a modified version of AMIE for link prediction.
	
	The results show that GRank is generally better than the other rule evaluation models because it considers the groundings of rules in great detail, whereas our method evaluates them indirectly.
	REE is competitive with GPro; its has worse results for WN18 and WN18RR and better results for FB15k and FB15k-237.
	{This shows REE can properly evaluate rules.}
	
\begin{table*}[tb]

	\caption{MRR and HITS@\textit{n} scores obtained for link prediction tasks with WN18, FB15k, WN18RR, and FB15k-237 datasets. 
		The highest result for each column is shown in bold.
		The results for GPro and GRank were reported by \citeauthor{ebisu-ichise-2019-graph} \shortcite{ebisu-ichise-2019-graph}, 
		those for ConvE were reported by \protect\citeauthor{DBLP:journals/corr/DettmersMSR17} \shortcite{DBLP:journals/corr/DettmersMSR17},
		and those for PRA were reported by \citeauthor{Liu:2016:HRW:2911451.2911509} \shortcite{Liu:2016:HRW:2911451.2911509}.
	}
\smallskip
		\scalebox{0.8}{
			\begin{tabular}{ccccccccccccccccc} \hline
				&\multicolumn{4}{c}{WN18}&\multicolumn{4}{c}{FB15k}&\multicolumn{4}{c}{WN18RR}&\multicolumn{4}{c}{FB15k-237}\\
				\cmidrule(rl){2-5}\cmidrule(rl){6-9}\cmidrule(rl){10-13}\cmidrule(rl){14-17}
				&\multicolumn{1}{c}{MRR}&\multicolumn{3}{c}{HITS@}&\multicolumn{1}{c}{MRR}&\multicolumn{3}{c}{HITS@}&\multicolumn{1}{c}{MRR}&\multicolumn{3}{c}{HITS@}&\multicolumn{1}{c}{MRR}&\multicolumn{3}{c}{HITS@}\\
				\cmidrule(rl){3-5}\cmidrule(rl){7-9}\cmidrule(rl){11-13}\cmidrule(rl){15-17}
				Model&&{1}&{3}&{10}&&{1}&{3}&{10}&&{1}&{3}&{10}&&{1}&{3}&{10}\\\hline 
				GPro&{0.950}&{0.946}&{0.954}&{0.959}&0.793&0.759&0.810&0.858&{0.467}&{0.430}&{0.485}&{0.543}&0.229&0.163&0.250&0.360\\
				GRank (fdMAP)&{0.950}&{0.946}&{0.954}&0.958&{0.842}&{0.816}&{0.856}&{0.891}&{0.470}&{0.437}&{0.482}&{0.539}&{0.322}&{0.239}&{0.352}&0.489\\
				REE&{0.942}&{0.940}&{0.944}&0.946&0.819&0.801&0.828&0.852&{0.437}&{0.403}&{0.452}&{0.504}&0.288&0.215&0.316&0.432\\\hline
				TorusE&0.951&0.947&0.954&0.960&0.810&0.768&0.835&0.884&0.477&0.439&0.490&0.551&0.346&0.252&0.380&0.535\\
				DistMult&0.922&0.891&0.952&0.956&0.840&0.802&0.865&0.906&0.460&0.416&0.472&0.548&0.354&0.260&0.389&0.543\\
				ComplEx&0.951 &0.945& 0.955 &0.962& 0.856& 0.827& 0.872& 0.909& 0.476& 0.429& 0.493& 0.564&　 0.365& 0.269& 0.401 &0.555\\
				ConvE&0.942&0.935&0.947&{0.955}&0.745&0.670&0.801&{0.873}&{0.46}&0.39&0.43&0.48&{0.316}&{0.239}&{0.350}&{0.491}\\
				PRA&0.458&0.422&--&0.481&0.336&0.303&--&0.392&--&--&--&--&--&--&--&--\\
				PBF with GRank&\textbf{0.953}&\textbf{0.948}&\textbf{0.956}&\textbf{0.963}&\textbf{0.870}&\textbf{0.845}&\textbf{0.882}&\textbf{0.915}&\textbf{0.494}&\textbf{0.453}&\textbf{0.509}&\textbf{0.576}&\textbf{0.376}&{0.282}&\textbf{0.413}&\textbf{0.564}\\
				PBF with REE&0.952&0.947&0.955&0.962&0.868&0.844&0.880&0.914&0.491&0.450&0.506&\textbf{0.576}&\textbf{0.376}&\textbf{0.284}&0.411&0.558\\\hline

			\end{tabular}
		}
	\label{Table 4}
\end{table*}
\subsubsection{Effect of Dimension}
It is known that a higher dimension for the embedding space improves the accuracy of embedding models.
The dimension may also affect the accuracy of rule evaluation.
Hence, we conducted the link prediction task using REE for various dimensions of the embedding space. The results are shown in Table \ref{dimension effect}.

The results show that a higher dimension for the embedding space improved the accuracy of rule evaluation.
This may explain why the dimension is important for link prediction: a space with higher dimension can more properly distinguish paths and learn rules. Hence, a model with a higher dimension can better predict links.

The results also show that the MRR score obtained with a relatively low dimension is close to the maximum score.
REE with a low-dimensional embedding space can thus be used when fast evaluation of rules is required.
\begin{table}[t]
	\caption{MRR scores obtained using REE for FB15k and FB15k-237 for various dimensions of the embedding space.}\smallskip
	\centering
	\resizebox{.95\columnwidth}{!}{
		\smallskip\begin{tabular}{clllll}\hline
			Dimension & 50 &100&500&1000&2000\\\hline
			FB15k&0.815&0.816&0.816&0.816&0.819\\
			FB15k-237&0.245&0.259&0.276&0.283& 0.288\\\hline
		\end{tabular}
	}
	\label{dimension effect}
\end{table}
\label{less dimension}
\subsection{Experimental Results for PBF}
{In this section, we discuss the results for PBF in terms of accuracy of link prediction to determine whether PBF can integrate different approaches.}
The results of the link prediction tasks for PBF are shown in Tables \ref{Table 4} with those of other models, where PBF employed GRank or REE to select paths.

The results show that ComplEx is really efficient. We can see the importance of the attention mechanism by comparing ComplEx with TorusE. The high accuracy of ComplEx is due to its ability to utilize a variety of rules and store information separately using the attention mechanism, as discussed and experimentally shown in previous sections.
PBE obtained the best results for all datasets when it incorporated ComplEx and observed feature models.
Especially, PBF improves the scores for HITS@1.
These results show that PBF compensates for the disadvantage of ComplEx, namely mixed and messy information, with carefully selected information.  

Particularly interesting observations are the results for PBF with REE.
This combination is really competitive with PBF with GRank even though it did not evaluate rules directly; instead, the rules were evaluated based on embeddings.
{The results suggest that REE is sufficient for PBF.}

\section{Conclusion}
\label{conclusions}
In this paper, we first unified state-of-the-art knowledge graph embedding models.
We generalized KGLG to AKGLG, where each entity and relation additionally has attention vectors, which are used to separately store information on a Lie group.
Then, the main embedding models were shown to be instances of AKGLG.
We proposed a method for evaluating rules based on the embeddings of AKGLG called REE.
Finally, we proposed a framework called PBF for incorporating AKGLG and observed feature models.
PBF compensates for the disadvantages of different approaches, which all utilize path information for link prediction.

We conducted experiments to evaluate the proposed methods.
REE was evaluated in terms of calculation time and link prediction accuracy.
The results showed that REE can reliably evaluate rules and that its calculation time is lower than that for traditional rule evaluation models for some standard datasets.
These results imply that existing models that are instances of AKGLG are really utilizing rules and we can understand what embeddings learned through REE .
PBF also outperformed existing models.
The results comprehensively show that AKGLG and observed feature models can effectively work together.

Embedding models can be further extended in the future.
However, we think we need always to take how a model deal with rules for further development into consideration.
That guarantees an extended model is interpretable and may be able to work with observed feature models more effectively.

\bibliographystyle{aaai}
\bibliography{bib2}

\begin{thebibliography}{}

\bibitem[\protect\citeauthoryear{Auer \bgroup et al\mbox.\egroup
  }{2007}]{DBLP:conf/semweb/AuerBKLCI07}
Auer, S.; Bizer, C.; Kobilarov, G.; Lehmann, J.; Cyganiak, R.; and Ives, Z.~G.
\newblock 2007.
\newblock {DB}pedia: {A} nucleus for a web of open data.
\newblock In {\em Proceedings of The Semantic Web, 6th International Semantic
  Web Conference},  722--735.

\bibitem[\protect\citeauthoryear{Bollacker \bgroup et al\mbox.\egroup
  }{2008}]{Bollacker:2008:FCC:1376616.1376746}
Bollacker, K.; Evans, C.; Paritosh, P.; Sturge, T.; and Taylor, J.
\newblock 2008.
\newblock Freebase: A collaboratively created graph database for structuring
  human knowledge.
\newblock In {\em Proceedings of the 2008 ACM SIGMOD International Conference
  on Management of Data},  1247--1250.

\bibitem[\protect\citeauthoryear{Bordes \bgroup et al\mbox.\egroup
  }{2013}]{DBLP:conf/nips/BordesUGWY13}
Bordes, A.; Usunier, N.; Garc{\'{\i}}a{-}Dur{\'{a}}n, A.; Weston, J.; and
  Yakhnenko, O.
\newblock 2013.
\newblock Translating embeddings for modeling multi-relational data.
\newblock In {\em Advances in Neural Information Processing Systems},
  2787--2795.

\bibitem[\protect\citeauthoryear{Bordes, Chopra, and Weston}{2014}]{D14-1067}
Bordes, A.; Chopra, S.; and Weston, J.
\newblock 2014.
\newblock Question answering with subgraph embeddings.
\newblock In {\em Proceedings of the 2014 Conference on Empirical Methods in
  Natural Language Processing},  615--620.

\bibitem[\protect\citeauthoryear{Daiber \bgroup et al\mbox.\egroup
  }{2013}]{Daiber:2013:IEA:2506182.2506198}
Daiber, J.; Jakob, M.; Hokamp, C.; and Mendes, P.~N.
\newblock 2013.
\newblock Improving efficiency and accuracy in multilingual entity extraction.
\newblock In {\em Proceedings of the 9th International Conference on Semantic
  Systems},  121--124.

\bibitem[\protect\citeauthoryear{Dettmers \bgroup et al\mbox.\egroup
  }{2018}]{DBLP:journals/corr/DettmersMSR17}
Dettmers, T.; Minervini, P.; Stenetorp, P.; and Riedel, S.
\newblock 2018.
\newblock Convolutional 2d knowledge graph embeddings.
\newblock In {\em Proceedings of the Thirtieth {AAAI} Conference on Artificial
  Intelligence}.

\bibitem[\protect\citeauthoryear{Ebisu and
  Ichise}{2018}]{DBLP:journals/corr/abs-1711-05435}
Ebisu, T., and Ichise, R.
\newblock 2018.
\newblock Toruse: Knowledge graph embedding on a lie group.
\newblock In {\em Proceedings of the Thirtieth {AAAI} Conference on Artificial
  Intelligence}.

\bibitem[\protect\citeauthoryear{Ebisu and Ichise}{2019a}]{KGLG}
Ebisu, T., and Ichise, R.
\newblock 2019a.
\newblock Generalized translation-based embedding of knowledge graph.
\newblock {\em IEEE Transactions on Knowledge and Data Engineering}.

\bibitem[\protect\citeauthoryear{Ebisu and
  Ichise}{2019b}]{ebisu-ichise-2019-graph}
Ebisu, T., and Ichise, R.
\newblock 2019b.
\newblock Graph pattern entity ranking model for knowledge graph completion.
\newblock In {\em Proceedings of the 2019 Conference of the North {A}merican
  Chapter of the Association for Computational Linguistics: Human Language
  Technologies},  988--997.

\bibitem[\protect\citeauthoryear{Gal{\'{a}}rraga \bgroup et al\mbox.\egroup
  }{2013}]{DBLP:conf/www/GalarragaTHS13}
Gal{\'{a}}rraga, L.~A.; Teflioudi, C.; Hose, K.; and Suchanek, F.~M.
\newblock 2013.
\newblock {AMIE:} association rule mining under incomplete evidence in
  ontological knowledge bases.
\newblock In {\em Proceedings of 22nd International World Wide Web Conference},
   413--422.

\bibitem[\protect\citeauthoryear{Gal{\'{a}}rraga \bgroup et al\mbox.\egroup
  }{2015}]{DBLP:journals/vldb/GalarragaTHS15}
Gal{\'{a}}rraga, L.; Teflioudi, C.; Hose, K.; and Suchanek, F.~M.
\newblock 2015.
\newblock Fast rule mining in ontological knowledge bases with {AMIE+}.
\newblock {\em {VLDB} J.} 24(6):707--730.

\bibitem[\protect\citeauthoryear{Guo \bgroup et al\mbox.\egroup
  }{2016}]{DBLP:conf/emnlp/GuoWWWG16}
Guo, S.; Wang, Q.; Wang, L.; Wang, B.; and Guo, L.
\newblock 2016.
\newblock Jointly embedding knowledge graphs and logical rules.
\newblock In {\em Proceedings of the 2016 Joint Conference on Empirical Methods
  in Natural Language Processing},  192--202.

\bibitem[\protect\citeauthoryear{Guo \bgroup et al\mbox.\egroup
  }{2018}]{DBLP:conf/aaai/GuoWWWG18}
Guo, S.; Wang, Q.; Wang, L.; Wang, B.; and Guo, L.
\newblock 2018.
\newblock Knowledge graph embedding with iterative guidance from soft rules.
\newblock In {\em Proceedings of the Thirty-Second AAAI Conference on
  Artificial Intelligence},  4816--4823.

\bibitem[\protect\citeauthoryear{Hakimov, Oto, and
  Dogdu}{2012}]{Hakimov:2012:NER:2237867.2237871}
Hakimov, S.; Oto, S.~A.; and Dogdu, E.
\newblock 2012.
\newblock Named entity recognition and disambiguation using linked data and
  graph-based centrality scoring.
\newblock In {\em Proceedings of the 4th International Workshop on Semantic Web
  Information Management},  1--7.

\bibitem[\protect\citeauthoryear{Ji \bgroup et al\mbox.\egroup
  }{2015}]{ji-etal-2015-knowledge}
Ji, G.; He, S.; Xu, L.; Liu, K.; and Zhao, J.
\newblock 2015.
\newblock Knowledge graph embedding via dynamic mapping matrix.
\newblock In {\em Proceedings of the 53rd Annual Meeting of the Association for
  Computational Linguistics},  687--696.

\bibitem[\protect\citeauthoryear{Kipf and
  Welling}{2016}]{DBLP:journals/corr/KipfW16}
Kipf, T.~N., and Welling, M.
\newblock 2016.
\newblock Semi-supervised classification with graph convolutional networks.
\newblock {\em CoRR} abs/1609.02907.

\bibitem[\protect\citeauthoryear{Lacroix, Usunier, and
  Obozinski}{2018}]{pmlr-v80-lacroix18a}
Lacroix, T.; Usunier, N.; and Obozinski, G.
\newblock 2018.
\newblock Canonical tensor decomposition for knowledge base completion.
\newblock In {\em Proceedings of the 35th International Conference on Machine
  Learning},  2863--2872.

\bibitem[\protect\citeauthoryear{Lao and Cohen}{2010}]{DBLP:journals/ml/LaoC10}
Lao, N., and Cohen, W.~W.
\newblock 2010.
\newblock Relational retrieval using a combination of path-constrained random
  walks.
\newblock {\em Machine Learning} 81(1):53--67.

\bibitem[\protect\citeauthoryear{Lao, Mitchell, and
  Cohen}{2011}]{DBLP:conf/emnlp/LaoMC11}
Lao, N.; Mitchell, T.~M.; and Cohen, W.~W.
\newblock 2011.
\newblock Random walk inference and learning in {A} large scale knowledge base.
\newblock In {\em Proceedings of the 2011 Conference on Empirical Methods in
  Natural Language Processing},  529--539.

\bibitem[\protect\citeauthoryear{Lin \bgroup et al\mbox.\egroup
  }{2015}]{DBLP:conf/aaai/LinLSLZ15}
Lin, Y.; Liu, Z.; Sun, M.; Liu, Y.; and Zhu, X.
\newblock 2015.
\newblock Learning entity and relation embeddings for knowledge graph
  completion.
\newblock In {\em Proceedings of the Twenty-Ninth {AAAI} Conference on
  Artificial Intelligence},  2181--2187.

\bibitem[\protect\citeauthoryear{Liu \bgroup et al\mbox.\egroup
  }{2016}]{Liu:2016:HRW:2911451.2911509}
Liu, Q.; Jiang, L.; Han, M.; Liu, Y.; and Qin, Z.
\newblock 2016.
\newblock Hierarchical random walk inference in knowledge graphs.
\newblock In {\em Proceedings of the 39th International ACM SIGIR Conference on
  Research and Development in Information Retrieval},  445--454.

\bibitem[\protect\citeauthoryear{Miller}{1995}]{Miller:1995:WLD:219717.219748}
Miller, G.~A.
\newblock 1995.
\newblock Wordnet: A lexical database for {E}nglish.
\newblock {\em Commun. ACM} 38(11):39--41.

\bibitem[\protect\citeauthoryear{Nickel, Tresp, and
  Kriegel}{2011}]{DBLP:conf/icml/NickelTK11}
Nickel, M.; Tresp, V.; and Kriegel, H.
\newblock 2011.
\newblock A three-way model for collective learning on multi-relational data.
\newblock In {\em Proceedings of the 28th International Conference on Machine
  Learning},  809--816.

\bibitem[\protect\citeauthoryear{Qian \bgroup et al\mbox.\egroup
  }{2018}]{ijcai2018-596}
Qian, W.; Fu, C.; Zhu, Y.; Cai, D.; and He, X.
\newblock 2018.
\newblock Translating embeddings for knowledge graph completion with relation
  attention mechanism.
\newblock In {\em Proceedings of the Twenty-Seventh International Joint
  Conference on Artificial Intelligence},  4286--4292.

\bibitem[\protect\citeauthoryear{Suchanek, Kasneci, and
  Weikum}{2007}]{DBLP:conf/www/SuchanekKW07}
Suchanek, F.~M.; Kasneci, G.; and Weikum, G.
\newblock 2007.
\newblock Yago: a core of semantic knowledge.
\newblock In {\em Proceedings of the 16th International Conference on World
  Wide Web},  697--706.

\bibitem[\protect\citeauthoryear{Toutanova and
  Chen}{2015}]{observed-versus-latent-features-for-knowledge-base-and-text-inference}
Toutanova, K., and Chen, D.
\newblock 2015.
\newblock Observed versus latent features for knowledge base and text
  inference.
\newblock In {\em Proceedings of the 3rd Workshop on Continuous Vector Space
  Models and Their Compositionality}.

\bibitem[\protect\citeauthoryear{Trouillon \bgroup et al\mbox.\egroup
  }{2016}]{DBLP:conf/icml/TrouillonWRGB16}
Trouillon, T.; Welbl, J.; Riedel, S.; Gaussier, {\'{E}}.; and Bouchard, G.
\newblock 2016.
\newblock Complex embeddings for simple link prediction.
\newblock In {\em Proceedings of the 33rd International Conference on Machine
  Learning},  2071--2080.

\bibitem[\protect\citeauthoryear{Wang \bgroup et al\mbox.\egroup
  }{2014}]{DBLP:conf/aaai/WangZFC14}
Wang, Z.; Zhang, J.; Feng, J.; and Chen, Z.
\newblock 2014.
\newblock Knowledge graph embedding by translating on hyperplanes.
\newblock In {\em Proceedings of the Twenty-Eighth {AAAI} Conference on
  Artificial Intelligence},  1112--1119.

\bibitem[\protect\citeauthoryear{Yang \bgroup et al\mbox.\egroup
  }{2014}]{DBLP:journals/corr/YangYHGD14a}
Yang, B.; Yih, W.; He, X.; Gao, J.; and Deng, L.
\newblock 2014.
\newblock Embedding entities and relations for learning and inference in
  knowledge bases.
\newblock {\em CoRR} abs/1412.6575.

\end{thebibliography}
\end{document}